\documentclass[onefignum,onetabnum]{siamonline190516}
\usepackage{amsmath, amssymb, mathrsfs}
\usepackage{lipsum}
\usepackage{amsfonts}
\usepackage{graphicx}
\usepackage{url}
\usepackage{multirow}
\usepackage{vcell}
\usepackage{xcolor,soul,framed} 

\colorlet{shadecolor}{yellow}

\usepackage{enumitem}
\setlist[enumerate]{leftmargin=.5in}
\setlist[itemize]{leftmargin=.5in}

\usepackage{algorithm}
\usepackage{algorithmic}

\newtheorem{thm}{Theorem}[section]

\newtheorem{rem}[thm]{Remark}

\graphicspath{{./figs/}}

\makeatletter
\DeclareFontFamily{U}{tipa}{}
\DeclareFontShape{U}{tipa}{m}{n}{<->tipa10}{}
\newcommand{\arc@char}{{\usefont{U}{tipa}{m}{n}\symbol{62}}}%

\newcommand{\arc}[1]{\mathpalette\arc@arc{#1}}

\newcommand{\arc@arc}[2]{%
  \sbox0{$\m@th#1#2$}%
  \vbox{
    \hbox{\resizebox{\wd0}{\height}{\arc@char}}
    \nointerlineskip
    \box0
  }%
}
\makeatother

\usepackage{pgf}

\headers{A Characteristic Function-based Algorithm for Geodesic Active Contours}{J. Ma, D. Wang, X.P. Wang, X.P. Yang}

\title{A Characteristic Function-based Algorithm for Geodesic Active Contours\thanks{Submitted to the editors \today.
\funding{Jun Ma and Xiaoping Yang acknowledge support by China's Ministry of Science and Technology (No. 2020YFA0713800) and the National Natural Science Foundation of China (No. 11971229, No. 12090023). Dong Wang acknowledges support by the University Development Fund from The Chinese University of Hong Kong, Shenzhen (UDF01001803). Xiao-Ping Wang acknowledges support by the Hong Kong Research Grants Council GRF grants 16303318 and 16305819.}}}

\author{Jun Ma\thanks{Department of Mathematics, Nanjing University of Science and Technology, Nanjing, 210094, P. R. China
  (\email{junma@njust.edu.cn}).}
\and Dong Wang\thanks{School of Science and Engineering, The Chinese University of Hong Kong, Shenzhen, Guangdong 518172, P. R. China
  (\email{wangdong@cuhk.edu.cn}). Corresponding author.}
\and  Xiao-Ping Wang\thanks{Department of Mathematics, Hong Kong University of Science and Technology, Clear Water Bay, Kowloon, Hong Kong, P. R. China
  (\email{mawang@ust.hk}.)}
  \and  Xiaoping Yang\thanks{Department of Mathematics, Nanjing University, Nanjing, 210093, P. R. China
  (\email{xpyang@nju.edu.cn}). Corresponding author.}} 

\usepackage{amsopn}

\begin{document}

\maketitle

\begin{abstract}
Active contour models have been widely used in image segmentation, and the level set method (LSM) is the most popular approach for solving the models, via implicitly representing the contour by a level set function. However, the LSM suffers from high computational burden and numerical instability, requiring additional regularization terms or re-initialization techniques.
In this paper, we use characteristic functions to implicitly represent the contours, propose a new representation to the geodesic active contours and derive an efficient algorithm termed as the iterative convolution-thresholding method (ICTM). Compared to the LSM, the ICTM is simpler and much more efficient. In addition, the ICTM enjoys most desired features of the level set-based methods. Extensive experiments, on 2D synthetic,  2D ultrasound, 3D CT, and 3D MR images for nodule, organ and lesion segmentation,  demonstrate that the proposed method not only obtains comparable or even better segmentation results (compared to the LSM) but also achieves significant acceleration.
\end{abstract}



\section{Introduction}

Active contours have been widely used in various segmentation tasks \cite{szeliski2010CVBook} and image modalities \cite{Xu2000SegReview}, such as organs \cite{ray2003lungSeg} in magnetic resonance (MR) scans, tumors in computed tomography (CT) scans \cite{GuiCT, GACTumorCT}, and ultrasound images \cite{GuiUltrasound, GACTumorUltra}. Basically, there are mainly two types of active contour models: edge-based active contours ({\it e.g.}, \cite{caselles1997GAC, li2010DRLSE, khadidos2017weightedGAC}) and region-based active contours ({\it e.g.}, \cite{chan2001CV, ismail2005multiregion, li2008RSF, lankton2008localizing}).
Edge-based active contours are driven by edge indicator functions that are commonly defined by image gradients. The contour evolution is expected to stop on boundaries with high gradient magnitude.
Region-based active contours are driven by regional information that can be defined by intensity statistical information inside and outside of the contour. The contour is expected to evolve to a position where the regional information inside and outside of the contour reaches a balance.

Edge-based active contours were firstly proposed by Kass et al. \cite{kass1988snakes} in 1988 (also termed as the snake model). The contour (explicitly represented by a parametric curve) evolves by the image gradient to the boundary of the desired object. When one only tracks the explicit curve, the snake model is very efficient and requires low memory, allowing a fast evolution of an accurate boundary. However, the snake model suffers from numerical instabilities, and is difficult to automatically handle topological changes of the curve during the evolution (typically, it works only for a single closed curve) \cite{CremersDiffSnake, li2010DRLSE}.

To improve the evolution of the contour of the snake model, Caselles et al. proposed geodesic active contours \cite{caselles1997GAC} in 1997. The contour is implicitly represented by a level set function that automatically allows topological changes including splitting and merging and simultaneous segmentation of single or multiple objects.
The key idea of the level set method (LSM), introduced by Osher and Sethian \cite{osher1988fronts}, is to represent a curve as the zero level set of a graph function defined in a higher dimensional space.
Nowadays, the LSM has been widely used in many applications including computer vision, computational geometry, fluid dynamics, material science, and so on (see \cite{osher2001LSReview} and references therein for more details). In particular, using the LSM to implicitly represent the contour and approximately solve the active contour models becomes the most popular choice \cite{Ismail2010LevelSetBook}.

Even though the geodesic active contour model (GAC) allows to change the topology of the curve during evolution, it still suffers from numerical instability. 
In fact, even if the analytical model could generate a correct contour for all times, it might happen, for either the analytical or the numerical reason, that the gradient of the level set function would become ``too small'' or ``too large'' on the contour ({\it i.e.}, the zero-level set). The gradient of a level set function being too small will result in the location of the zero level set (the interface) being sensitive to perturbation. If the gradient is too large, one loses accuracy in the contour representation. To avoid this problem, the level set function is periodically reinitialized as a distance function from the interface, allowing to keep the norm of the gradient close to the unity and avoiding ill-conditioning. However, the reinitialization procedure usually involves many tricks, for example, it is hard to decide when it should be applied. Li et al. \cite{li2010DRLSE} proposed a penalty term to keep the regularity of the level set function during evolution. The core idea is to use the intrinsic property of the signed distance function: the magnitude of the gradient of the signed distance function equals one. A penalty term is then introduced to penalize this constraint. In addition, they proposed to keep the magnitude of the gradient of the level set function to be $1$ in a neighborhood of the zero level set and the value of the level set function to be a constant at locations far away from the zero level set, to accelerate the algorithm in practical implementation.
Compared to the classical snake model, GAC has achieved significant improvements by using the level set function to represent the contour. However, it is still inefficient because the level set evolution is evolved by a time-dependent partial differential equation that has large computational burden to obtain a solution.
Several acceleration algorithms have been designed for the GAC \cite{goldenberg2001fastGAC,appleton2005globallyGAC,bresson2007fastGAC}, but solutions are usually obtained by explicit schemes that may have stability issues regarding the choice of the time step.

Moreover, geodesic active contours should couple with an area term to speed up the curve motion when the initial contour is far away from the desired object boundaries. However, the sign of the area term weight is required to be manually determined. For example, when the level set function takes negative values inside the zero contour and positive values outside, the area weight should be positive if the initial contour is placed outside the object. By contrast, the area weight should be negative if the initial contour is placed inside the object.
To ease this problem, hybrid active contours \cite{zhang2008HAC, xu2014hybridAC, liu2014medical, zhang2019RESLS, HAC-TIP2020} introduce region descriptors \cite{chan2001CV, li2008RSF} that can allow the contour adaptively evolve to the object boundary without manually specifying the sign of the weight. This is because the region descriptors are derived from image features (e.g., intensities). For example, the region term in Chan-Vese model~\cite{chan2001CV} assumes the image has piecewise constant intensities. The region term in~\cite{li2011MR} aims to find the multiplicative components of a give image: a smooth bias filed function and a piecewise constant function.
Hybrid active contours have wide applications, such as retinal images~\cite{zhao2015HybridAC}, liver vessel~\cite{zeng2018HAC-Vessel}, myocardium images~\cite{huang2018hybrid} and so on, but they still suffer from low computational efficiency and and usually have many hyper-parameters.

In addition to level set-based curve evolution, characteristic function can also be used for curve representation and evolution. The idea was firstly introduced by Merriman, Bence, and Osher (MBO) \cite{MBO1992,merriman1994motion} for the modeling and simulation of motion by mean curvature, which iteratively diffuses the characteristic function of the interior region of the curve followed by thresholding. The MBO scheme has been adapted for piecewise constant ``Mumford-Shah" style image segmentation \cite{tai2005MBOremark,MBO06MSModel} and segmentation problems on general weighted graphs (see \cite{MBO13Graph, bertozzi2016diffuse,van2014mean} for examples). More recently, Esedoglu and Otto~\cite{esedog2015Threshold} proposed a new formulation to interpret the MBO scheme as a minimizing movement scheme of a Lyapunov functional of characteristic functions, which can be directly generalized to multi-phase mean curvature motions with arbitrary surface tensions.   The method has attracted much attention due to its simplicity and unconditional stability.  It has subsequently been extended to many problems, including the problem of area or volume preserving interface motion \cite{ruuth2003simple,jacobsauction},  wetting dynamics \cite{xu2016efficient,Wang_2019,wang2019efficient,jiang2018efficient}, target-valued harmonic maps \cite{osting2017generalized,wang2019interface,wang2018diffusion,Osting_2020},  high-order geometric motions \cite{Esedoglu_2008}, and so on. 

\begin{table*}[!htbp]
\caption{Features of three contour representation methods. $\otimes$ means that this feature requires special design for the method ({\it e.g.}, additional penalty term).}\label{tab:features}
\centering
\resizebox{\textwidth}{!}{
\begin{tabular}{llccc}
\hline
Representation Type   & Formulation & Computational Efficiency   & Adaptively Topological Change &  Stability      \\ \hline
Parametric Curve       & $C(s,t):[0,1)\times [0,\inf)\rightarrow \mathbf{R}^2$        &       $\surd$      &  $\times$                   &    $\times$     \\
Level Set              & $\{x\in \Omega\subset \mathbf{R}^2 | \phi(x,t)=0\}$       &       $\times$     & $\surd$                     &    $\otimes$    \\
Characteristic Function& $ u(\mathbf{x}) = \begin{cases}  1 \quad \text{if} \ \  \mathbf{x}\in \Omega \\ 0 \quad \text{otherwise}  \end{cases} $      &       $\surd$      & $\surd$                     &   $\surd$        \\ \hline
\end{tabular}}
\end{table*}


Characteristic function-related length regularizers have been widely used in existing segmentation models. In~\cite{pham2001spatialFC}, Pham proposed to add a membership function-based spatial penalty term to the fuzzy C-means objective function, which can introduce spatial smoothness. Moreover, centroidal Voronoi tessellation model~\cite{du2006CVT} was also proposed for image segmentation where its basic form is known as the K-means clustering. Furthermore, the basic CVT model was extended to edge-weighted centroidal Voronoi tessellation (EWCVT) clustering~\cite{wang2009edgeCVT}, where the pixels with its neighbors belong to different clusters are penalized and the approximate length regularization term is
the same as that in~\cite{pham2001spatialFC}. The EWCVT model was further improved by introducing image edge information and local variations of intensities, which can deal with inhomogeneous image segmentation \cite{liuJun2011PRfast,wang2011TIP-LocalVarCVT}.
In~\cite{chan2006GlobalMin,lie2006binaryLS}, Chan-Vese model was formulated in terms of binary characteristic functions and the curve length was expressed as the total variation of the characteristic function. The resulting optimizing algorithm~\cite{chan2006GlobalMin} can find a global minimizer for fixed values of the intensity cluster means.
In addition, the characteristic function-based regularizer can also be coupled with Gaussian mixture model (GMM), which can add spatial smoothness constraint to the classical GMM~\cite{liu2013GMM}. 

Recently, motivated by Esedoglu and Otto's new formulation~\cite{esedog2015Threshold}, Wang et al. \cite{wang2017JCP, wang2019ICTM} proposed to use a characteristic function to represent the contour in region-based active contour models, where the perimeter of the contour is approximated by a heat kernel convolution with a characteristic function. Furthermore, they derived an iterative convolution-thresholding method (ICTM) to minimize a general energy functional with general fidelity terms, which enjoys the unconditionally energy-decaying property. Numerical experiments in \cite{wang2017JCP, wang2019ICTM} have shown that the ICTM is simple, efficient, and applicable to a wide range of region-based segmentation models. Furthermore, the iteration algorithm can be unfolded as part of the deep neural network~\cite{liuJun2020RegCNN}, which can impose spatial regularization in an end-to-end way.

Most of the above segmentation methods focus on obtaining global minimizer under various situations, especially for the images with noise and inhomogeneous intensities, and the energy functionals of the length regularizers are often spatially inhomogeneous. However, for the geodesic active contour, a local minimizer is desired rather than a global minimizer, which is common in many segmentation applications. Moreover, its energy functional is spatially inhomogeneous because the edge indicator function in this functional is inhomogeneous in the image domain.
Motivated by the MBO scheme \cite{MBO1992, MBO13Graph} and the new perimeter approximation formulation~\cite{esedog2015Threshold} and its wide extensions  \cite{wang2017JCP, wang2019ICTM, wang2020efficient, liuJun2020RegCNN}, in this paper, we propose to use a characteristic function to implicitly represent the evolving contour, approximate the geodesic active contour energy functional and derive an efficient algorithm to minimize the energy. The main contributions are summarized as follows:
\begin{itemize}
  \item[(1)] An iterative convolution-thresholding method is developed to solve the GAC. Experiments on synthetic, ultrasound, CT and MR images demonstrate that the ICTM is more efficient than the classical level set method, which is the most popular algorithm in this area.
  \item[(2)] The energy-decaying property of the proposed algorithm is theoretically proved.
\end{itemize}

The paper is organized as follows. In Section~\ref{s:method} we introduce the characteristic function-based representation of the contour, give an approximation of the energy functional and derive an efficient algorithm to minimize the energy. In Section~\ref{s:exp} extensive experiments are performed to verify the efficiency of the proposed method.  We discuss the intuitive understanding of the ICTM and other potential applications in Section~\ref{s:dis} and draw some conclusions in Section~\ref{s:con}.

\section{Preliminaries}
For an image $I$ on a domain $\Omega$, there are usually three methods to represent the evolution of object boundary curves in it: 1) parametric curves, 2) level set functions, and 3) characteristic functions. Table \ref{tab:features} presents the main features of the above three representation methods. In the following, we first review the parametric curve- and level set-based approaches.


\subsection{Parametric curve representation}
Let $C(q):[0,1]\rightarrow R^2$ be a parametric curve. To find the object boundary, the classical snakes model \cite{kass1988snakes} defines the following energy functional associated with the curve $C$
\begin{equation}
\begin{aligned}
    E(C) = & E_{internal} + E_{external} \\
    =&  \alpha\int_0^1 |C'(q)|^2dq + \beta\int_0^1 |C''(q)|^2 \ dq 
   - \lambda\int_0^1|\nabla I(C(q))|\ dq,
\end{aligned}
\end{equation}
where $\alpha, \beta$, and $\lambda$ are real positive constants.
The first two terms belong to the internal energy that controls the smoothness of the contours, while the third term is the external energy that drives the contours towards the boundary of the object.

The snakes model is a pioneer that formulates the image segmentation problem as an energy functional minimization problem. Using a parametrized planar curve to represent an object contour allows a fast evolution with a spline function method \cite{CremersDiffSnake}. However, it suffers from the fixed topological property. For example, if there are more than one objects in a given image and the initial segmentation contour surrounds the objects, the snake model can not detect all objects. In other words, the classical snakes model can not directly deal with topological changes.

\subsection{Level set representation}
To address the drawback of the snakes model, Caselles et al. \cite{caselles1997GAC} proposed the well-known GAC that is a geodesic computational problem in a Riemannian space whose metric is defined by the image information. The energy functional is represented by embedding the dynamic contour $C(s,t)$ as the zero level set of a time dependent level set function $\phi: \Omega\times[0,\infty]\rightarrow\mathbf{R}$
\begin{equation}
    \min\limits_\phi \int_\Omega g\delta(\phi)|\nabla\phi|\ d\mathbf{x},
\end{equation}
where $\delta$ is the Dirac delta function on the set $\phi = 0$, and $g:[0,+\infty)\rightarrow R^{+}$ is an edge indicator function. In general, $g$ is defined by
\begin{equation}\label{eq:g}
    g :=\frac{1}{1+|\nabla G_\sigma * I|^2},
\end{equation}
where $G_\sigma$ is a Gaussian kernel with a standard deviation $\sigma$ that is used to smooth the image. It is easy to see that $g$ takes smaller values on the object boundary where the gradient magnitude is larger.

One of the most significant advantages of level set-based contour representation is that it can handle topological changes ({\it e.g.}, merging and splitting) in a natural way, which is not allowed in parametric-based contour representation.
In practice, an area term is usually introduced to speed up the motion of the zero level set during the evolution, which is important when the initial contour is far away from the desired object boundaries. The Dirac delta function $\delta$ is approximated by $\delta_\epsilon$, which is defined by
\begin{equation}\label{eq:smoothdelta}
    \delta_\epsilon(\mathbf{x}) =
    \begin{cases}
    \frac{1}{2\epsilon} [1+\cos(\frac{\pi \mathbf{x}}{\epsilon})], \quad |\mathbf{x}|\leq \epsilon \\
    0, \quad |\mathbf{x}|> \epsilon
    \end{cases},
\end{equation}
and $\epsilon>0$ is a hyper-parameter that controls the band width of the non-zero region.
The energy functional is then defined by
\begin{equation}\label{eq:GACLS}
    E(\phi) = \alpha \int_\Omega g \delta_\epsilon(\phi)|\nabla\phi|d\mathbf{x} + \lambda \int_\Omega gH_\epsilon(-\phi)d\mathbf{x},
\end{equation}
where $H_\epsilon$ is the smoothed Heaviside function.

Although the level set methods have the desired property on handling topology changes, their applications suffer from issues on numerical instability. To be specific, a level set function is usually defined as a signed distance function and typically develops irregularities due to numerical errors during evolution, which could destroy the stability of the level set evolution.

\subsection{Distance regularized level set}
To address the numerical instability problem, reinitialization was periodically used in earlier level set methods \cite{chan2001CV} to force the level set to be a signed distance function during the contour evolution. To avoid the above problem, Li et al. \cite{li2010DRLSE} proposed the following distance regularized level set evolution (DRLSE) method
\begin{equation}\label{eq:DRLSE}
    \begin{aligned}
     E(\phi) =& \alpha \int_\Omega g \delta(\phi)|\nabla\phi|d\mathbf{x} + \lambda \int_\Omega gH(-\phi)d\mathbf{x}
      + \mu\int_\Omega p(|\nabla \phi|) \ d\mathbf{x},
    \end{aligned}
\end{equation}
where $\mu>0$ is a weight hyper parameter and $p(x)$ is a potential function that is used to keep the signed distance regularity of the level set function. Typically, the potential function $p(x)$ can be defined as a single-well potential
\begin{equation}
    p(x) := \frac{1}{2} (x-1)^2,
\end{equation}
or a double-well potential
\begin{equation}
    p(x) :=
    \begin{cases}
    \frac{1}{(2\pi)^2}(1-\cos(2\pi x)) \quad \text{if}\quad x \leq 1, \\
    \frac{1}{2} (x-1)^2 \quad \text{if} \quad x > 1.
    \end{cases}
\end{equation}
In general, double-well potential is the default setting because it is more robust than single-well potential.
With the smoothed Dirac delta function $\delta_\epsilon$ and Heaviside function $H_\epsilon$, one can derive the gradient flow of the energy functional (\ref{eq:DRLSE}) as
\begin{equation}\label{eq:GD-DRLSE}
    \frac{\partial \phi}{\partial t} = \alpha \delta_\epsilon(\phi) \text{div}(g\frac{\nabla \phi}{|\nabla \phi|}) + \lambda g\delta_\epsilon (\phi) + \mu \text{div}(d_p(|\nabla\phi|)\nabla\phi)
\end{equation}
where $d_p(x)=\frac{p'(x)}{x}$.

Although the LSM has more advantages than the classical parametric curve-based methods and the distance regularization approach and hybrid active contour model make the LSM more stable, they still have two main drawbacks: many hyper-parameters and low computational efficiency, because solving the level set-based GAC needs to update the level set function according to a partial differential equation (\ref{eq:GD-DRLSE}), which has large computational burden.
In the next section, we will present a characteristic function-based method for GAC, which significantly reduces the number of hyper-parameter and improve the computational efficiency.

\section{Proposed method}\label{s:method}
\subsection{Geodesic active contours with the characteristic function}\label{sec:characteristic}
A characteristic function can be used to represent the contour, which is defined by
\begin{equation}
  u(\mathbf{x}) =
  \begin{cases}
    1 \quad \text{if} \ \  \mathbf{x}\in \Omega_\Gamma \\
    0 \quad \text{otherwise}
  \end{cases},
\end{equation}
where $\Gamma$ is the object boundary and $\Omega_\Gamma$ denotes the region inside $\Gamma$.
It provides an alternative way to implicitly represent a curve, which not only owns the adaptively topological change property of the LSM but also is more computationally efficient and stable. In this paper, we propose to formulate the energy functional of GAC beyond existing level set methods. In particular, a characteristic function is introduced to approximate the energy (\ref{eq:GACLS}), which allows us to design a more efficient algorithm for GAC.

As shown in \cite{Miranda_2007}, a general boundary integral can be approximated using the characteristic functions $u$ by:
\begin{equation}
    \int_\Gamma g \ ds \approx \lim\limits_{\tau \rightarrow 0 }\sqrt{\frac{\pi}{\tau}}\int_{\mathbb R^n} g u G_\tau * (1-u) \ d\mathbf{x},
\end{equation}
or
\begin{equation}
    \int_\Gamma g \ ds \approx \lim\limits_{\tau \rightarrow 0 }\sqrt{\frac{\pi}{\tau}}\int_{\mathbb R^n} g (1-u)G_\tau * u\ d\mathbf{x}
\end{equation}
where $\tau$ is a free parameter, $*$ represents convolution between two functions, and
\begin{align*}
G_{\tau}(\mathbf{x}) = \frac{1}{(4 \pi \tau)^{n/2}}\exp(-\frac{|\mathbf{x}|^2}{4\tau}).
\end{align*}
Here $n$ is the dimension of the Euclidean space and $\Gamma$ could be an interface when $n=2$ or a surface when $n=3$. Esedoglu and Otto \cite{esedog2015Threshold} established a novel framework on modelling and simulating the multiphase mean curvature flow with arbitrary surface tensions based on this approximation, via a relaxation and linearization procedure.

To keep the symmetry of the formulation with respect to $u$ and $1-u$, combining with the area term, the geodesic active contour energy functional (\ref{eq:GACLS}) is approximated by\footnote{$\lambda$ in \eqref{eq:approxGACLS} is the same as $\lambda / \alpha$ in \eqref{eq:GACLS}.}
\begin{equation} \label{eq:approxGACLS}
\begin{aligned}
E^\tau(u):=& \sqrt{\frac{\pi}{\tau}}\int_{\Omega} \sqrt{g} u G_\tau*(\sqrt{g}(1-u)) + \lambda g u \ d\mathbf{x}.
\end{aligned}
\end{equation}

The convergence of \eqref{eq:approxGACLS} as $\tau \searrow 0$ is referred to \cite{Huweithesis}. One can also use above formula to design efficient algorithms for surface reconstruction from point clouds \cite{wang2020efficient}.

Then, the original GAC is approximated by the following energy functional minimization problem
\begin{equation}\label{eq:uGAC}
    u^* = \arg\min\limits_{u\in B} E^\tau (u),
\end{equation}
where
\begin{equation*}
    B:= \{u\in BV(\Omega, R) | u=\{0, 1\}\}
\end{equation*}
and $BV(\Omega, R)$ denotes the space of functions with bounded variation.


\subsection{Algorithm for problem (\ref{eq:uGAC})}
It is easy to see that the feasible set $B$ of the energy functional minimization problem (\ref{eq:uGAC}) is non-convex. To address this problem, we relax $B$ to its convex hull
\begin{equation*}
    K:=\{u\in BV(\Omega, R) | u\in[0,1]\}
\end{equation*}
 and derive the following relaxed minimization problem
\begin{equation}\label{eq:uGAC-relax}
    u^* = \arg\min\limits_{u\in K} E^\tau (u).
\end{equation}
Furthermore, we prove the equivalence between (\ref{eq:uGAC}) and (\ref{eq:uGAC-relax}) in the following lemma.

\begin{lemma}
The original problem (\ref{eq:uGAC}) is equivalent to the relaxed problem (\ref{eq:uGAC-relax}), in the sense that if $u^*$ is a solution of (\ref{eq:uGAC}), then it is also a solution of (\ref{eq:uGAC-relax}) and vice versa.
\end{lemma}

\begin{proof}
On the one hand, let $\hat{u}=\arg\min\limits_{u\in B}E^\tau(u)$, we have
\begin{equation}\label{eq:lemmaEq1}
    E^\tau(\hat{u}) = \min\limits_{u\in B} E^\tau(u).
\end{equation}
Then, it is obvious that
\begin{equation*}
    \arg \min\limits_{u\in B} E^\tau (u) \in K,
\end{equation*}
and
\begin{equation}
    E^\tau(\hat{u})\geq\min_{u\in K}E^\tau (u),
\end{equation}
because $B \subsetneqq K$.

On the other hand, let $\tilde{u}=\arg\min\limits_{u\in K}E^\tau(u)$, we can use reduction to absurdity to prove
\begin{equation*}
    \tilde{u} = \arg \min\limits_{u\in K} E^\tau (u) \in B.
\end{equation*}
Assume it is not true, then there exists a set $A\subseteq \Omega$ with nonzero measure and $a>0$ such that the minimizer $u^*$ satisfies
\begin{equation*}
    u^*(x) \in (a, 1-a), \forall x\in A.
\end{equation*}
Let $u^t = u^* + t\chi_A$ where $\chi_A$ is the characteristic function of $A$, we have $u^t\in K$ for any $|t|<a$.
Directly computing the first and the second derivatives of
\begin{equation*}
E^\tau(u^t) =\sqrt{\frac{\pi}{\tau}}\int_\Omega \sqrt{g}(u^* + t\chi_A)G_\tau * (\sqrt{g}(1-u^* - t\chi_A)) + \lambda g(u^* + t\chi_A) d\mathbf{x}
\end{equation*}
with respect to $t$, we have
\begin{equation*}
\frac{dE^\tau(u^t)}{dt} =  \sqrt{\frac{\pi}{\tau}}\int_\Omega \sqrt{g}\chi_A G_\tau * (\sqrt{g}(1-u^*-t\chi_A)) + \sqrt{g}(u^*+t\chi_A)G_\tau*(-\sqrt{g}\chi_A) + \lambda g\chi_A \ d\mathbf{x}
\end{equation*}
and
\begin{equation*}
    \frac{d^2E^\tau(u^t)}{dt^2} = -2\sqrt{\frac{\pi}{\tau}}\int_\Omega \sqrt{g}\chi_A G_\tau * (\sqrt{g}\chi_A) \ d\mathbf{x}.
\end{equation*}
Due to $\sqrt{g}\geq 0$ and $\sqrt{g}=0$ only on a set with zero measure, we have $\frac{d^2E^\tau(u^t)}{dt^2}<0$ especially at $t=0$ ({\it i.e.}, $u^*$). However, this result contradicts with the assumption that $u^*$ is a minimizer ($\frac{d^2E^\tau(u^t)}{dt^2}\geq0$ at $u^*$).

Thus, we have $\tilde{u}\in B$, and then
\begin{equation}\label{eq:lemmaEq4}
    E^\tau(\tilde{u}) = \min\limits_{u\in K}E^\tau(u)\geq E^\tau(\hat{u}).
\end{equation}
Finally, based on (\ref{eq:lemmaEq1})-(\ref{eq:lemmaEq4}), we obtain
\begin{equation}
    E^\tau(\hat{u}) = \min\limits_{u\in B}E^\tau(u) = \min\limits_{u\in K}E^\tau (u) = E^\tau(\tilde{u}).
\end{equation}

\end{proof}

Next, we derive an iterative method to solve the relaxed problem (\ref{eq:uGAC-relax}). It is easy to verify that $E^\tau (u)$ is a concave functional. Using the fact that the graph of the functional $E^\tau (u)$ is always below its linear approximation, we minimize the linearized functional of $E^\tau (u)$ based on the sequential linear programming because the minimizer of the linearized functional can give a smaller value in $E^\tau (u)$. The similar idea can also be found in the well-known difference of convex algorithm \cite{tao1997convex,tao1998dc}.
Specifically, at the $k$-th iteration $u^k$, we compute the linearization (the first order Taylor expansion) of $E^\tau(u)$ at $u^k$ as
\begin{equation}
    L^\tau(u, u^k) = \sqrt{\frac{\pi}{\tau}} \int_{\Omega} u\varphi^k d\mathbf{x},
\end{equation}
where
\begin{equation}
    \varphi^k = \sqrt{g} G_\tau * (\sqrt{g}(1-2u^k)) + \lambda g.
\end{equation}
Then, one obtains the $k+1$-th iteration $u^{k+1}$ by solving the following linearized problem:
\begin{equation}\label{eq:linear}
    u^{k+1} = \arg\min\limits_{u\in K}L^\tau(u,u^k).
\end{equation}
which can be solved a pointwise manner. That is, $\forall \mathbf{x}\in\Omega$, we solve 
\begin{equation}
    u^{k+1}(\mathbf{x}) = \arg \min\limits_{u(\mathbf{x})\in[0,1])} u(\mathbf{x})\varphi^k(\mathbf{x}).
\end{equation}
Because of the fact that the minimizer of a linear functional over a convex set must be reached at the boundary, we get
\begin{equation}
  u^{k+1}(\mathbf{x}) =
  \begin{cases}
    1 \quad \text{if}\ \  \varphi(\mathbf{x}) \leq 0 \\
    0 \quad \text{otherwise}
  \end{cases}.
\end{equation}

We summarize the proposed iterative convolution-thresholding method (ICTM) in Algorithm \ref{alg:GAC-TDM}.
\begin{algorithm}[!htbp]
\caption{The iterative convolution-thresholding method (ICTM) for geodesic active contours}
\label{alg:GAC-TDM}
\begin{algorithmic}
\REQUIRE Image edge indicator function $g$, $\tau>0$, $\lambda$, and initialization $u^0 \in B$.
\ENSURE Segmentation results $u^*\in B$;
\WHILE{not converged}
\STATE (1) Convolution. Fix $u^k$, compute $$\varphi^k (x) = \sqrt{g} G_\tau * (\sqrt{g}(1-2u^k)) + \lambda g$$
\STATE (2) Thresholding. Set $$
  u^{k+1}(\mathbf{x}) =
  \begin{cases}
    1 \quad \text{if}\ \ \varphi(\mathbf{x}) \leq 0 \\
    0 \quad \text{otherwise}
  \end{cases}.
$$
\ENDWHILE
\end{algorithmic}
\end{algorithm}

\begin{rem}
For the convergence criteria, we stop the iteration if
\begin{equation}
    \int_\Omega |u^{k+1}-u^k| \ d\mathbf{x} < tol
\end{equation}
where $tol$ is a given error tolerance ($1\times 10^{-5}$ in this paper).
\end{rem}

\begin{rem}
Compared to the LSM, the ICTM enjoys following advantages:
\begin{itemize}
    \item[(1)] No requirement for the additional regularization term: The LSM in (\ref{eq:DRLSE}) needs an auxiliary regularization term to maintain the numerical stability, while our algorithm is intrinsic stable during iterations as demonstrated in the following section.
    \item[(2)] Fewer hyper-parameters: The LSM has three model hyper-parameters\footnote{The standard deviation $\sigma$ in the Gaussian kernel of edge indicator function (Eq. (\ref{eq:g})) is excluded because this hyper-parameter is not always necessary. For example, if an input image is clean, we do not need to use Gaussian filter to smooth the image.}: $\alpha, \lambda$, and $\mu$ (can be reduced to two by some normalizations) and two algorithm hyper-parameters: time step and band width $\epsilon$ in $\delta_\epsilon$ and $H_\epsilon$, while our method reduces the number of hyper-parameters to one model parameter $\lambda$ and one joint model-algorithm hyper-parameter $\tau$.
    \item[(3)] Less computational burden: The LSM needs to solve a partial differential equation for the evolution of the contour as shown in the gradient flow (\ref{eq:GD-DRLSE}), while our algorithm only alternates simple convolution and thresholding operations.
\end{itemize}
\end{rem}

\begin{rem}
The GAC model can only deal with images whose contours are clear. One can also combine the ICTM for the GAC and the ICTM for other region based models (as those in \cite{wang2019ICTM}) for general images. 
\end{rem}

\subsection{Stability analysis}
In this section, we prove that the Algorithm~\ref{alg:GAC-TDM} is unconditionally stable for any $\tau>0$, which means the total energy $E^\tau(u)$ is decreasing during the iteration. Thus, the proposed method can always converge to a stationary configuration.

\begin{thm}\label{thm:decay}
  Let $u^k$ $(k=0,1,2,\cdots)$ be the $k$-th iteration derived in Algorithm \ref{alg:GAC-TDM}. We have
  \begin{equation*}
      E^{\tau}(u^{k+1}) \leq E^{\tau}(u^{k})
  \end{equation*}
  for any $\tau>0$.
\end{thm}
\begin{proof}
As for $E^\tau(u)$ defined in \eqref{eq:approxGACLS}, the linearization of $E^\tau(u)$ at $u^k$ is defined by:
$$L^\tau(u,u^k)  = \sqrt{\frac{\pi}{\tau}} \int_{\mathbb R^n} \sqrt{g} u G_\tau * (\sqrt{g} (1-2u^k)) + \lambda u g \ d\mathbf{x}.$$
Direct calculation yields that
\begin{equation*}
    \begin{aligned}
       E^\tau(u^k) &= \sqrt{\frac{\pi}{\tau}}\int_{\Omega} \sqrt{g} u^k G_\tau * \left(\sqrt{g} (1-u^k)\right) + \lambda gu^k d\mathbf{x}\\
        &= L^\tau(u^k,u^k)+ \sqrt{\frac{\pi}{\tau}}\int_{\Omega} \sqrt{g} u^k G_\tau * \left(\sqrt{g} u^k\right)\ d\mathbf{x}
    \end{aligned}
\end{equation*}
and
\begin{equation*}
\begin{aligned}
& E^\tau(u^{k+1})  = \sqrt{\frac{\pi}{\tau}}\int_{\Omega} \sqrt{g} u^{k+1} G_\tau * \left(\sqrt{g} (1-u^{k+1})\right) + \lambda gu^{k+1} d\mathbf{x}\\
& = L^\tau(u^{k+1},u^k) + 2\sqrt{\frac{\pi}{\tau}}\int_{\Omega} \sqrt{g}  u^{k+1} G_\tau * \left(\sqrt{g} u^k\right) \ d \mathbf{x} - \sqrt{\frac{\pi}{\tau}}\int_{\Omega} \sqrt{g}  u^{k+1} G_\tau * \left(\sqrt{g} u^{k+1}\right) \ d \mathbf{x}
\end{aligned}
\end{equation*}
Because $u^{k+1}$ is the solution from the sequential linear programming, we have $L^\tau(u^{k+1},u^k) \leq L^\tau(u^k,u^k)$. Then, we compute
\begin{align*}
E^\tau(u^{k+1})  - E^\tau(u^k) = L^\tau(u^{k+1},u^k) - L^\tau(u^k,u^k) + \mathcal L
\end{align*}
where \begin{align*}
\mathcal L = & - \sqrt{\frac{\pi}{\tau}}\int_{\Omega}  \sqrt{g}  (u^{k+1}-u^k)  G_\tau * \left(\sqrt{g}  (u^{k+1}-u^k) \right) \ d \mathbf{x}.
\end{align*}
Based on the semi-group property of the heat kernel convolution, {\it i.e.},
\[\int_{\mathbb R^n} f G_{\tau} * g  \ d \mathbf{x} = \int_{\mathbb R^n} (G_{\tau/2} *f) (G_{\tau/2} * g)  \ d \mathbf{x},\]
we have
\begin{align*}
\mathcal L=& - \sqrt{\frac{\pi}{\tau}} \int_{\mathbb R^n}  \left[G_{\tau/2} * \left(\sqrt{g}(u^{k+1}-u^k) \right) \right]^2 \ d \mathbf{x}  \leq 0 .\end{align*}
Therefore, we are led that
$E^\tau(u^{k+1})  - E^\tau(u^k) \leq 0$ because $L^\tau(u^{k+1},u^k) - L^\tau(u^k,u^k) \leq 0$ from the derivation of the algorithm.
\end{proof}

\begin{rem}
The proposed ICTM is motivated by the general idea of the MBO scheme: characteristic functions are introduced to implicitly describe the interface motion by mean curvature, which can be used for the contour evolution in segmentation tasks. However, different from MBO schemes \cite{MBO1992,esedog2015Threshold} or extensions on weighted graphs \cite{MBO13Graph}, we consider the case involving a spatially inhomogeneous interface motion ({\it i.e.},  iterations of the active contour) where the inhomogeneity comes from the inhomogeneous edge indicator function $g(x)$ in the domain. To handle such inhomogeneity, we proposed a symmetric, quadratic, concave, and inhomogeneous approximation approach \eqref{eq:approxGACLS} using Gaussian kernels. It then allows us to design efficient iteration algorithm ({\it i.e.}, Algorithm \ref{alg:GAC-TDM}) and prove its unconditional stability ({\it i.e.}, Theorem \ref{thm:decay}).
\end{rem}

\section{Numerical experiments}
\label{s:exp}
In this section, we present four groups of image segmentation experiments to demonstrate that compared with the level set method (LSM), the proposed iterative convolution-thresholding method (ICTM) can achieve comparable or even better segmentation results but with significantly fewer iterations.
Specifically, we first evaluate the LSM and the ICTM on two synthetic images, which show that the ICTM can also adaptively handle the topological changes, such as merging and splitting.
Then, we show the performance of Algorithm~\ref{alg:GAC-TDM} on three different types of images, including ultrasound (2D), CT (3D) and MR (3D) images, and compare with the well-known LSM-based GAC \cite{li2010DRLSE} (DRLSE) to show the efficiency of the proposed method. The software code of LSM is obtained from Chunming Li's homepage\footnote{https://www.engr.uconn.edu/~cmli/\label{ChunmingCode}}.

For fair comparison, we also implement our algorithm with pure MATLAB code\footnote{https://www.mathworks.com/matlabcentral/fileexchange/91875-ictm-gac-segmentation}. All the experiments are ran on a Windows 10 laptop with Intel(R) Core(TM) i7-6700HQ CPU @ 2.60GHz, 24 GM RAM.
Our code will be publicly available upon acceptance.
In all experiment, if not specifically pointed out, we set $\sigma$ in \eqref{eq:g} to be 15 and $\epsilon$ in \eqref{eq:smoothdelta} to be 1.5.

\begin{rem}
One can also use banded LSM or ICTM to reduce the computational complexity at each iteration. For the fair comparison, we simply perform the LSM and the ICTM in the whole computational domain.
\end{rem}


\subsection{Experiments on synthetic images}\label{sec:A}
We first apply the proposed method on two synthetic images to show that using the ICTM can automatically handle the topological changes, such as merging and splitting, during evolution, which is the most important feature in level set-based methods.
The hyper-parameters of the LSM follow the default setting in Chunming Li's code\textsuperscript{\ref{ChunmingCode}}.
As for Algorithm~\ref{alg:GAC-TDM}, we set $\tau=2$ and choose the parameter $\lambda$ as -0.3 and 0.3 for merging and splitting experiments, respectively.

\begin{figure*}[!htbp]
\centering
\includegraphics[scale=0.7]{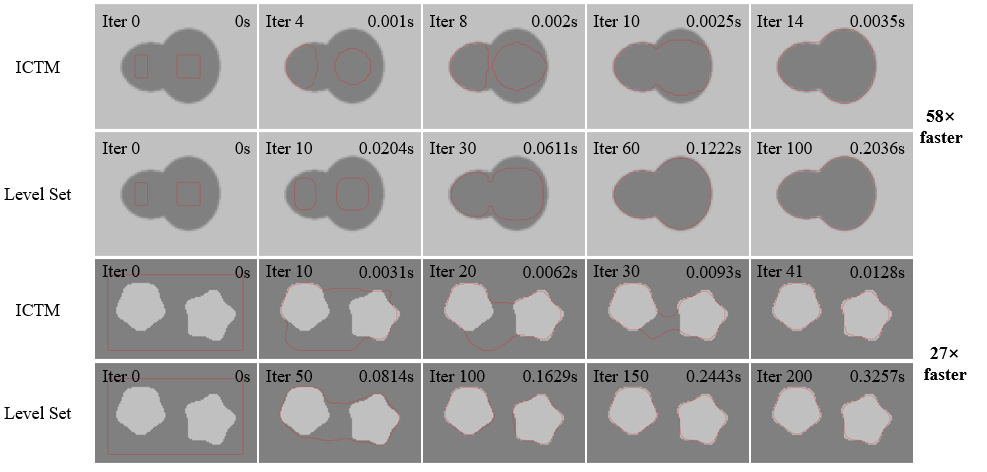}
\caption{Experimental results on two synthetic images. {\bf First two rows:} selected snapshots of the merging process. {\bf Last two rows:} selected snapshots of the splitting process. {\bf The red rectangles} in the first column are the contour initialization. At the top of each image, the corresponding iteration number and running time are listed. See Section~\ref{sec:A}.}\label{fig:exp-demo}
\end{figure*}

Figure \ref{fig:exp-demo} displays the segmentation results of the proposed ICTM and the LSM on two synthetic images. The first two rows show the merging process while the last two rows show the splitting process, using selected snapshots during the iteration. The corresponding number of iterations and running time are listed at the top of each image.
Based on these segmentation results, we observe that
\begin{itemize}
    \item[(1)] Both the ICTM and the LSM can adaptively handle the topological changes. Specifically, the contours can adaptively merge and split during iterations.
    \item[(2)] Both the ICTM and the LSM can achieve same (or similar) segmentation results on the two synthetic images because they are just different approximations to the same model ({\it i.e.}, GAC).
    \item[(3)] Compared to the LSM, the ICTM achieves the same segmentation results with many fewer iterations and much less running time. In particular, ICTM is approximate 58 times and 27 times faster than the LSM on the contour merging and splitting results, respectively.
\end{itemize}

\subsection{Breast nodule segmentation in ultrasound images}
\label{ss:exp-breast}
To validate the performance and efficiency of our method on real images, we apply it to breast nodule segmentation in ultrasound images. In this experiment, we use the BUSI dataset~\cite{BUSIDataset} and select 100 random benign breast ultrasound images to compare the efficiency between the LSM and the ICTM. The same rectangle initialization is generated based on ground truth for images. The width and height of each rectangle is half of the major axis and minor axis of the ground truth's bounding box, respectively.
For a fair comparison, we apply grid search to each method to find the best set of hyper-parameters that can achieve the best average Dice similarity coefficient (DSC) on this dataset. The main contribution in this work is a more efficient method. Thus we try our best to achieve the best performance for both LSM and ICTM, and then compare their running efficiency.
Specifically, we set $\alpha=5$, $\lambda=-3$, $\mu=0.2$ and $\Delta t = 1$ for the LSM, respectively.
For ICTM, we set $\tau=2$ and $\lambda=-0.2$, respectively.

\begin{figure*}[ht!]
\centering
\includegraphics[scale=0.6]{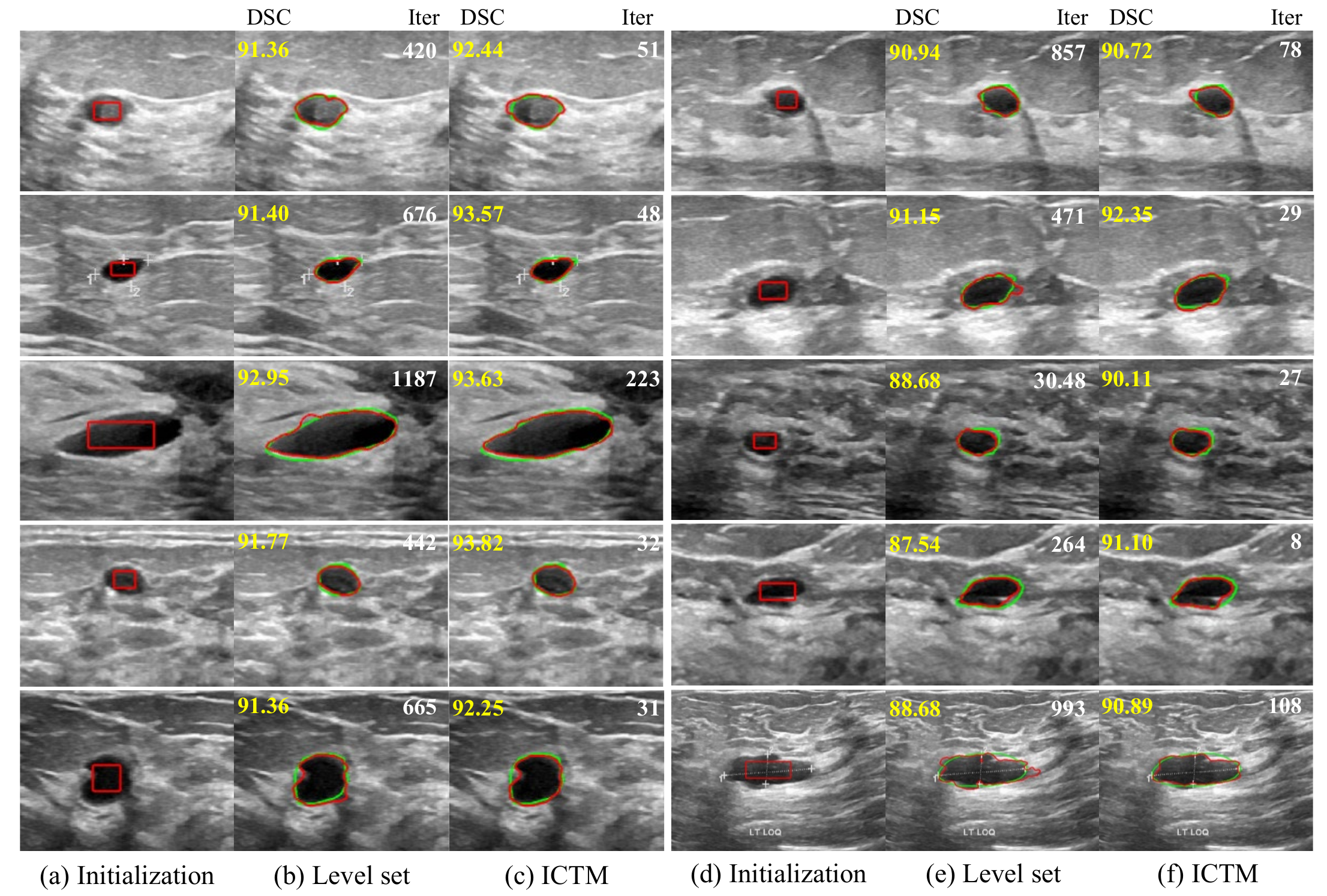}
\caption{Breast nodule ultrasound image segmentation results. {\bf Columns (a)} and {\bf (d)} are input images and corresponding initialization (red rectangles). {\bf Columns (b)} and {\bf (e)} are segmentation results of the LSM. {\bf Columns (c)} and {\bf (f)} are segmentation results of Algorithm~\ref{alg:GAC-TDM}. {\bf Green} and {\bf red} contours denote ground truths and segmentation results, respectively. {\bf Yellow} and {\bf white} numbers at the top of images indicate segmentation accuracy (Dice similarity coefficient (DSC)) and the number of iterations. See Section~\ref{ss:exp-breast}.}\label{fig:exp-breast}
\end{figure*}

Table \ref{tab:exps} shows the quantitative results of breast ultrasound image segmentation. The proposed ICTM method is slightly better than the LSM with 1.35\% improvements in average DSC.
The number of average iteration of our ICTM is 58, which is 12.4 $\times$ fewer than the LSM.
For the average running time, the LSM takes about 74 seconds per image while the proposed ICTM only needs 0.66 seconds per image, achieving more than 100 times acceleration.
Figure \ref{fig:exp-breast} displays 10 random selected images with segmentation results by the LSM and our ICTM, with corresponding DSC (yellow) and running time (white) printed on.
The two methods start with the same initialization, and we observe that
\begin{itemize}
    \item The ICTM achieves similar or even better segmentation results than the LSM, indicating the feasibility of applying the ICTM into real image segmentation.
    \item The ICTM requires much less running time than the LSM, implying the high efficiency.
\end{itemize}

\begin{table*}[ht!]
\caption{Quantitative results in different datasets in terms of average Dice coefficient similarity (DSC), running time and the number of iterations (\# of Iter.). The arrows indicate which direction is better. Bold numbers mean that the improvements are statistically significant at $p<0.01$. See Sections~\ref{ss:exp-breast}, \ref{sec:19ct}, and \ref{sec:mr}.}\label{tab:exps}
\centering
\resizebox{\textwidth}{!}{
\begin{tabular}{cccccc}
\hline
Dataset     & Method      & DSC (\%) $\uparrow$       & Time (s) $\downarrow$    & \# of Iter. $\downarrow$ & Acceleration (in Iter.)   \\ 
\hline
Breast      & Level set   & 87.75 $\pm$ 5.34          & 74.01 $\pm$ 44.18  & 719 $\pm$ 398                & 12.4 $\times$  \\ 
\cline{2-5}
Ultrasound  & ICTM (ours) & \textbf{89.10 $\pm$ 3.99} & \textbf{0.66 $\pm$ 0.61}    & \textbf{58 $\pm$ 48}         & faster         \\ 
\hline
COVID-19    & Level set   & 92.18 $\pm$ 2.35          & 1956.6 $\pm$ 550.2 & 7826 $\pm$ 2200              & 3.86 $\times$  \\ 
\cline{2-5}
CT          & ICTM (ours) & 92.35 $\pm$ 3.21          & \textbf{69.2 $\pm$ 29.3}    & \textbf{1384 $\pm$ 586}      & faster         \\ 
\hline
Liver Tumor & Level set   & 85.79 $\pm$ 6.32          & 104.5 $\pm$ 83.2   & 475 $\pm$ 378                & 5.65 $\times$  \\ 
\cline{2-5}
MR          & ICTM (ours) & 86.44 $\pm$ 6.61          & \textbf{1.52 $\pm$ 1.78}    & \textbf{123 $\pm$ 145}       & faster         \\
\hline
\end{tabular}}
\end{table*}

\subsection{Lung segmentation in COVID-19 CT} \label{sec:19ct}
To validate the effectiveness of the ICTM on 3D organ segmentation tasks, we apply it to lung segmentation in COVID-19 CT scans.
We use the public COVID-19-CT-Seg dataset \cite{COVID-19-SegBenchmark} and select 10 earlier COVID-19 lung CT scans for segmentation experiments. The image size of each slice is $512\times512$, and the number of image slices ranges from 40 to 400.
We set the same initialization for both the ICTM and the LSM, which is generated by eroding the ground truth with a sphere structure element (radius$=$10).

\begin{figure*}[!htbp]
\centering
\includegraphics[scale=0.6]{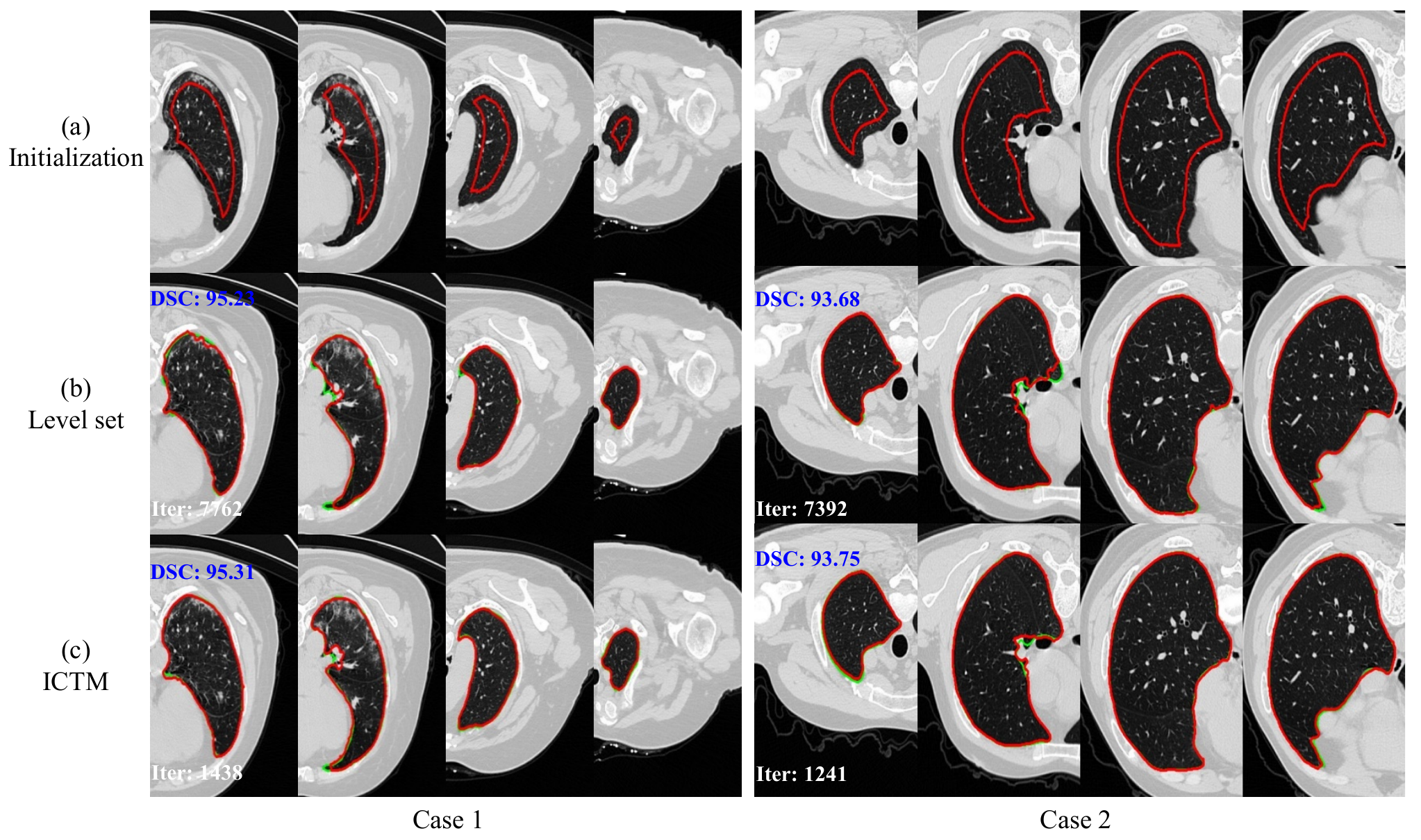}
\caption{Zoomed examples for lung CT segmentation results. {\bf Row (a)} : input images and corresponding initialization (red rectangles). {\bf Row (b)} : segmentation results of the LSM. {\bf Row (c)} : segmentation results of the proposed ICTM. {\bf Green} and {\bf red} contours denote ground truths and segmentation results, respectively. {\bf Blue} and {\bf white} numbers at the top of images point out segmentation accuracy (DSC) and the number of iterations. See Section~\ref{sec:19ct}.}\label{fig:exp-lung}
\end{figure*}

For fair comparison, we apply the grid search to tune the hyper-parameters, which aims to achieve the best performance for both LSM and ICTM. Quantitative and qualitative segmentation results are displayed in Table~\ref{tab:exps} and Figure~\ref{fig:exp-lung}, respectively. We observe that both two methods achieve comparable average DSC without significant differences, which implies that the ICTM can be an alternative choice (beyond the LSM) for the GAC. 
The average iterations of ICTM is 1384, which is 3.86 $\times$ lower than the LSM. 
As for the efficiency, the LSM spends averagely 1956.6 seconds for each case while the ICTM only needs 69.2 seconds, which achieves about $28.3$ times acceleration.

\begin{figure}[!htbp]
\centering
\includegraphics[scale=0.4]{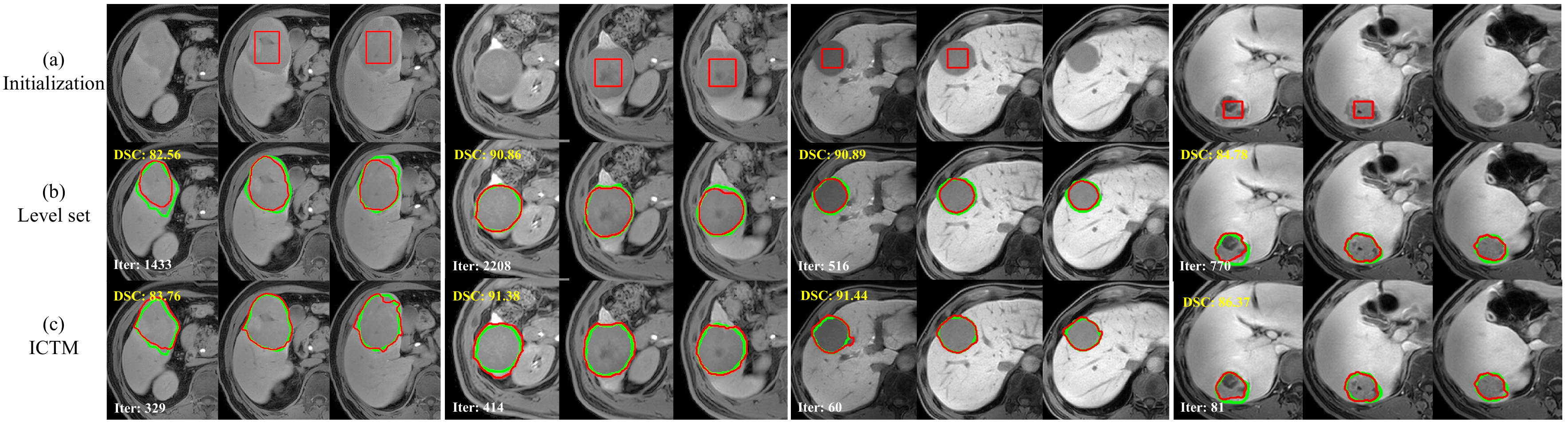}
\caption{Zoomed examples of liver lesion MR image segmentation results. Green and red contours denote ground truths and segmentation results, respectively. Blue and white numbers at the top of images point out segmentation accuracy (DSC) and the number of iterations. Some images do not have initializations because the segmentation method is applied in a 3D manner and not all tumor slices need initializations.
See Section~\ref{sec:mr}.}\label{fig:exp-liver}
\end{figure}

\subsection{Liver lesion segmentation in MR}\label{sec:mr}
To validate the effectiveness of our ICTM on 3D lesion segmentation tasks, we apply it to liver lesion segmentation in liver MR scans. We randomly collect 20 liver MR scans from a local hospital. Three experienced radiologists manually annotate them, and majority vote is used to generate final labels. The image sizes range from $256\times256\times105$ to $400\times400\times120$.
The initialization of each MR scan is a cuboid inside the tumor. Figure \ref{fig:exp-liver} (a)  shows some initialization results (red rectangle) in 2D slices. It should be noted that not all tumor slices have initializations such as the images in the first and forth columns in Figure~\ref{fig:exp-liver} (a) because the segmentation method is applied in a 3D manner. Thus, we do not need to give initializations in each slice. In fact, only half of the tumor slices have initializations.

For fair comparison, we also apply the grid search to tune the hyper-parameters. Quantitative and qualitative segmentation results are displayed in Table~\ref{tab:exps} and Figure~\ref{fig:exp-liver}, respectively. We observe that, again, both two methods achieve similar average DSC without significant differences, indicating that the ICTM can obtain similar results to the LSM for the GAC.
However, the ICTM requires fewer iterations compared to the LSM, which is about $5.65$ times acceleration, implying the high efficiency of the ICTM.

\section{Discussion}
\label{s:dis}
In Section~\ref{s:exp}, we have applied the proposed ICTM on synthetic, ultrasound, CT and MR images to show its effectiveness on nodule, organ and lesion segmentation. Compared to the LSM, the ICTM obtains similar or even better results but achieves dozens or hundreds of times faster execution times. Furthermore, we discuss the intuitive understanding on the advantages of the ICTM ({\it e.g.}, efficiency) and many potential applications especially in the modern deep learning era.

\subsection{Why the proposed ICTM is faster than the LSM?}
This is contributed to the simple and inherent features (Table~\ref{tab:features}) of using characteristic functions to implicitly represent a contour. Specifically, there are several main reasons:
\begin{itemize}
    \item[(1)] Each iteration in the ICTM (Algorithm~\ref{alg:GAC-TDM}) is much simpler than each iteration in the LSM as shown in (\ref{eq:GD-DRLSE}).
    \item[(2)] The ICTM directly minimizes the geodesic active contour energy functional, while the LSM usually needs to minimize additional energy term to stabilize the iteration.
    \item[(3)] At each iteration, the ICTM can find the optimal minimizer of the linearized functional. This is because the optimal minimizer of a linear functional over a convex set can be reached at the boundary. Moreover, the minimizer can give a smaller value in $E^\tau$ because the graph of the functional $E^\tau$ (concave) is always below its linear approximation. This accelerates the convergence of the ICTM. In the LSM, one needs to solve the level set-based partial differential equation (PDE) with a relatively small time step. This step more or less restricts the decay of the energy (at least not optimal). What's worse, the reinitialization step (or adding penalty terms in the level set equation) usually increases the energy, which decreases the value of the energy minimized at each iteration (increases the value of $E^\tau$). This makes the LSM converge slower. In the ICTM, thanks to the concavity of $E^\tau$, the minimizer at each iteration automatically gives a new partition ({\it i.e.}, the minimizer automatically remains at characteristic functions). No reinitialization and related regularization techniques are needed in the ICTM. We plot the energy curves of both methods during iterations for the synthetic image segmentation (see Figure~\ref{fig:exp-demo}), and observe that ICTM converges faster than the LSM.  In particular, we observe more oscillations in the energy curve for the LSM.
\end{itemize}

\begin{figure}[t!]
\centering
\includegraphics[scale=0.45]{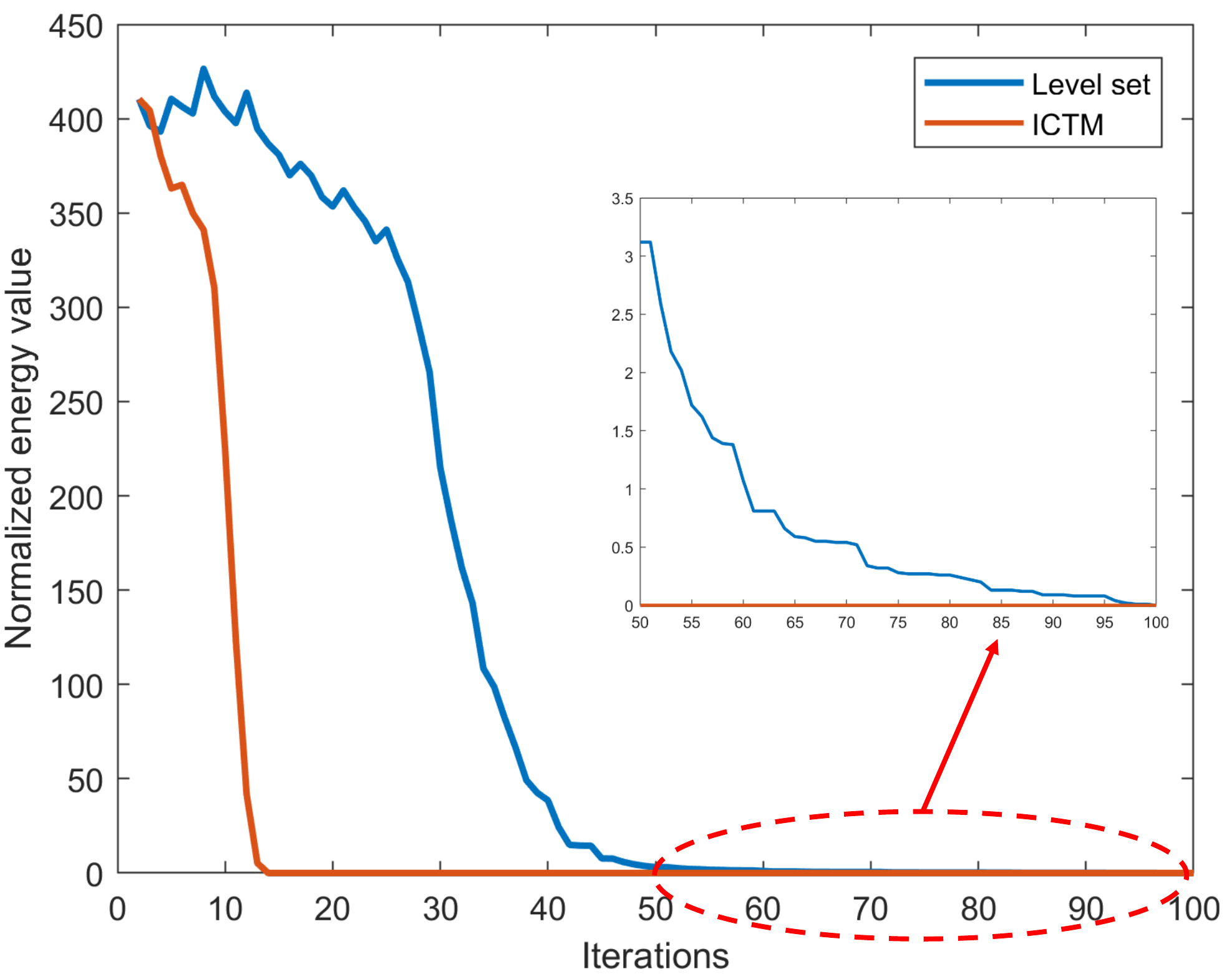}
\caption{Energy curves during iterations of the LSM and the ICTM in the synthetic image segmentation. For better normalization, we apply max-min normalization to the energy values. We zoom in the energy curve of the last 50 iterations to better show the difference between two methods. It can be found that ICTM converges faster than the LSM. In particular, more oscillations occur in the energy curve for the LSM.}\label{fig:energy}
\end{figure}

\subsection{What kind of images can the proposed ICTM work?}
The ICTM is a method for approximately solving an image segmentation model ({\it i.e.}, minimizing an objective functional).
On one hand, from extensive numerical experiments in this paper, we claim that, the ICTM is much more efficient than the LSM for various segmentation tasks.
On the other hand, the proposed ICTM may not be outside of the application scope of the LSM. This is because both two methods aim to approximately solve the same GAC (\ref{eq:GACLS}), which mainly determines the accuracy of the segmentation.

\subsection{What is the role of the traditional model-based geodesic active contours in modern deep learning era?}
Although deep learning-based segmentation methods have been increasingly popular and dominating current segmentation tasks, these methods generally require much annotated training data that is difficult to obtain in medical images. Deep learning techniques are still open for many mathematical explanations and theories, which may cause incomprehensible segmentation results.

Model-based GAC has interpretable nature and still plays important roles in following three circumstances:
\begin{itemize}
    \item[(1)] Assisting radiologists to annotate medical images ({\it e.g.}; tumor segmentation \cite{TumorSeg}).
    \item[(2)] Serving as a post-processing method to refine the segmentation results that are generated by deep learning-based approaches ({\it e.g.}; breast tumor \cite{CNNACBreast}, liver \cite{CNNACLiver}, and dental root \cite{ma2019CBCT} segmentation).
    \item[(3)] Explicitly embedding shape information ({\it e.g.}; a left ventricle shape model that is learned by auto-encoder network can be embedded into the GAC \cite{MIADeepLS}).
    \item[(4)] Reformulating the active contour model as a loss function~\cite{LossOdyssey} to guide CNNs to learning richer features, such as Mumford--Shah loss \cite{LossMumford}, level set loss \cite{LossLS}, active contour loss \cite{LossAC}, and geodesic active contour loss~\cite{LGAC-TMI}.
\end{itemize}

The proposed ICTM is expected to be applied to some of these situations directly, obtaining a dramatic acceleration. We leave these applications as our future work and will be reported elsewhere.

\section{Conclusion and future work}
\label{s:con}
In this paper, we proposed an efficient iterative convolution-thresholding method (ICTM) to solve the wildly used geodesic active contours (GAC). The method mainly relies on a characteristic function-based representation for the contour and an integral approximation of the energy functional. A relaxation and linearization approach is used to derive the ICTM method. Extensive numerical experiments on four different types of images are presented to show the performance of the proposed method, indicating a dramatic improvement in the efficiency (compared to the level set-based approaches).

In the future, we will create an image segmentation benchmark specialized for the model-based segmentation community. Although deep learning has achieved state of the arts in many image segmentation tasks~\cite{minaee2021TPAMISegSurvey, ma2021SOTASeg}, it also provides new challenges and opportunities for model-based segmentation methods. Specifically, given initial (inaccurate) segmentation results generated by cutting-edge deep learning solutions, how or what kind of model-based methods can consistently improve the segmentation accuracy? We will create an image segmentation dataset with challenging images, especially for the images that deep learning fails to segment, and include the popular segmentation models as baselines, which can enable fair comparisons.

\section*{Acknowledgment}
The authors highly appreciate Jian He for providing approved liver MR scans and annotations used in the experiments. The authors would also like to thank Ziwei Nie and Yiming Gao for their valuable discussions.

\bibliographystyle{siamplain}
\bibliography{refs}

\end{document}